\documentclass[colorinlistoftodos]{article}

\usepackage[preprint]{nips_2018}
\usepackage[utf8]{inputenc} 
\usepackage[T1]{fontenc}    
\usepackage{hyperref}       
\usepackage{url}            
\usepackage{booktabs}       
\usepackage{amsfonts}       
\usepackage{nicefrac}       
\usepackage{microtype}      
\usepackage{wrapfig}
\newcommand{\rulesep}{\unskip\ \vrule\ }

\usepackage{pdfpages}
\usepackage{float}
\usepackage{tikz}
\usetikzlibrary{positioning}
\usetikzlibrary{calc}

\usepackage{amssymb,amsmath}
\usepackage{mathtools}

\usepackage{color}
\usepackage[font=small,labelfont=bf]{caption}
\usepackage{algorithm,algorithmicx,algpseudocode}
\algnewcommand\algorithmicinput{\textbf{Input:}}
\algnewcommand\INPUT{\item[\algorithmicinput]}
\algnewcommand\algorithmicoutput{\textbf{Output:}}
\algnewcommand\OUTPUT{\item[\algorithmicoutput]}
\algnewcommand\algorithmicidea{\textbf{Idea:}}
\algnewcommand\IDEA{\item[\algorithmicidea]}
\algnewcommand\algorithmicinit{\textbf{Initialize:}}
\algnewcommand\INIT{\item[\algorithmicinit]}

\usepackage{amsthm}
\newtheorem{assumption}{Assumption}

\usepackage{fancyvrb}
\usepackage{subfig}

\theoremstyle{definition}

\usepackage{wrapfig}
\usepackage{lipsum}  

\usepackage{todonotes}

\usepackage[font=small,labelfont=bf]{caption}

\title{Semantic RL with Action Grammars: Data-Efficient Learning of Hierarchical Task Abstractions}

\author{
  Robert Tjarko Lange \\
  Einstein Center for Neurosciences Berlin\\
  Technical University Berlin\\
  \url{robert.lange17@imperial.ac.uk} \\
	\and 
  A. Aldo Faisal \\
  Imperial College London \\
  Department of Computing \& Bioengineering\\
  \url{a.faisal@imperial.ac.uk} \\
}

\begin{document}

\maketitle

\begin{abstract}
Hierarchical Reinforcement Learning algorithms have successfully been applied to temporal credit assignment problems with sparse reward signals. However, state-of-the-art algorithms require manual specification of sub-task structures, a sample inefficient exploration phase or lack semantic interpretability. Humans, on the other hand, efficiently detect hierarchical sub-structures induced by their surroundings.
It has been argued that this inference process universally applies to language, logical reasoning as well as motor control. 
Therefore, we propose a cognitive-inspired Reinforcement Learning architecture which uses grammar induction to identify sub-goal policies. By treating an on-policy trajectory as a sentence sampled from the policy-conditioned language of the environment, we identify hierarchical constituents with the help of unsupervised grammatical inference. The resulting set of temporal abstractions is called \textit{action grammar} \citep{Pastra_2012} and unifies symbolic and connectionist approaches to Reinforcement Learning. It can be used to facilitate efficient imitation, transfer and online learning.
\end{abstract}

\section{Introduction}\label{ch1:intro}
\begin{figure}[H]
    \centering
    \includegraphics[width=0.525\linewidth]{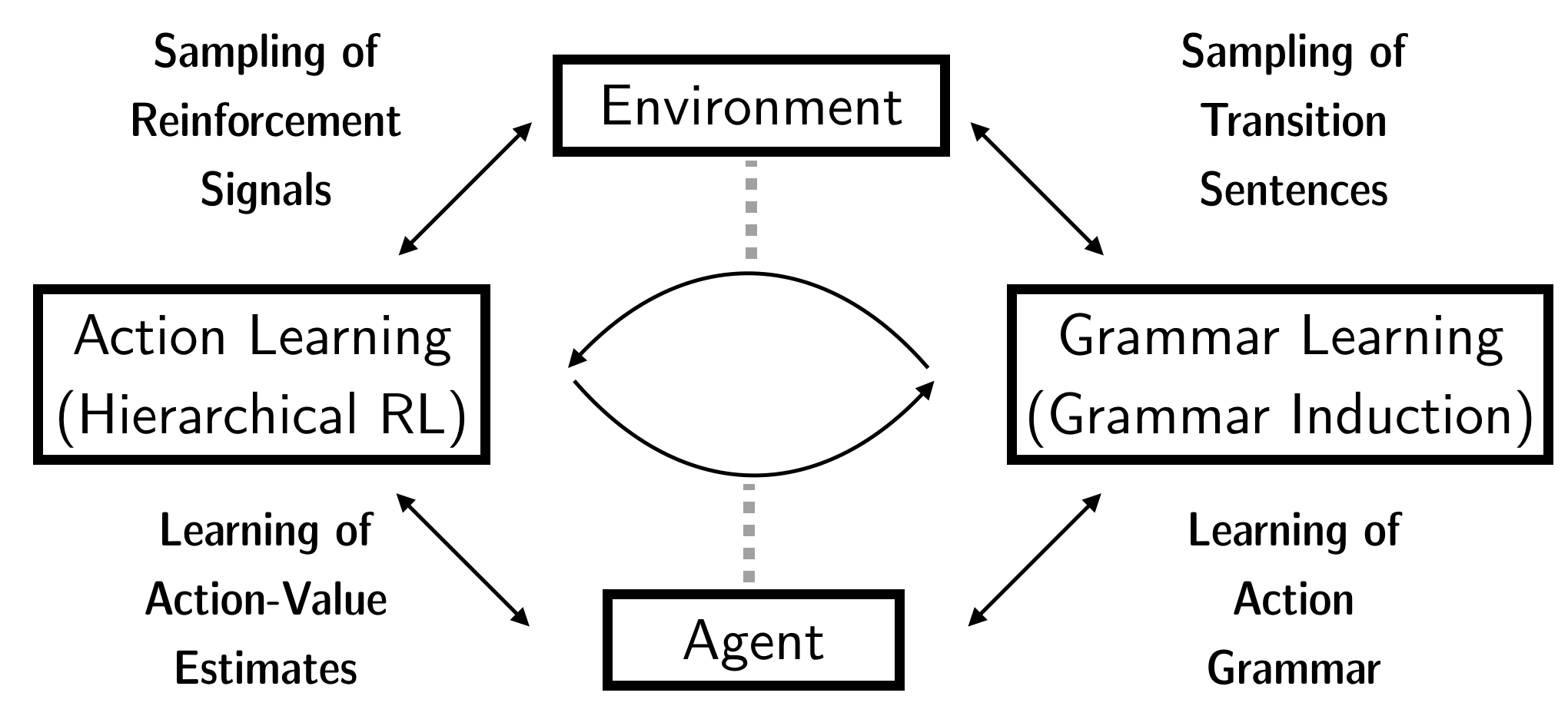}
    \rulesep
    \includegraphics[width=0.45\linewidth]{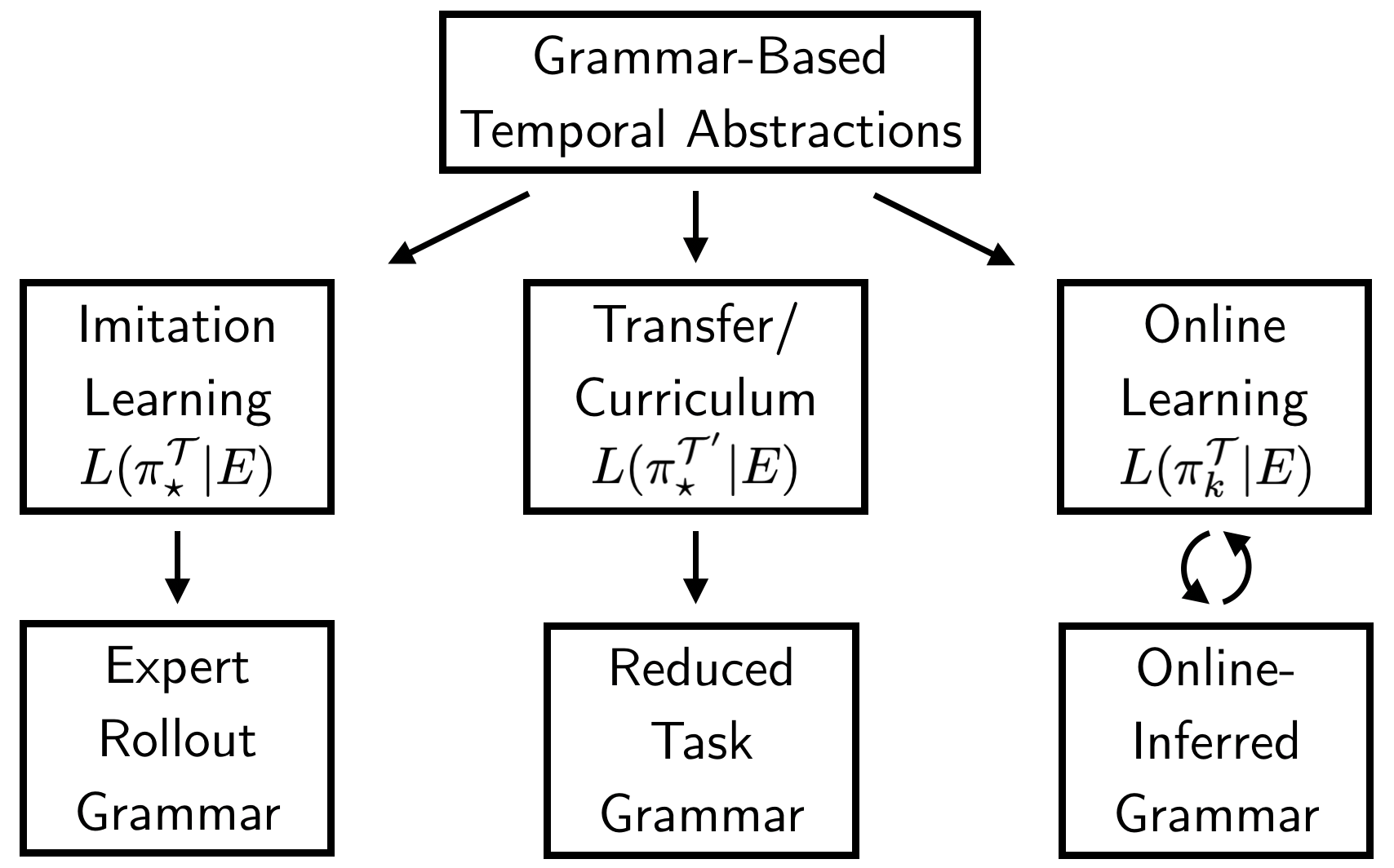}
    \caption{\textbf{Left.} Action grammars closed alternation loop. \textbf{Right.} Action grammar applications.}
    \label{fig:loop_ag}
\end{figure}

Human inductive biases enable us to rapidly infer hierarchical rule-based structures ('grammars') from language, visual input as as well as auditory stimuli \cite{Frank_2009, Marcus_2007}. Several neuroimaging studies provide evidence for a universal process of hierarchical grammar comprehension in the brain \cite{Frank_2018, Brennan_2016, Nelson_2017} that extends to human motor control of real-world tasks \cite{Pastra_2012, Stout_2018}. According to these works we hypothesise that by processing action trajectories of an expert, a student is able to efficiently learn policies over higher level sequences of task abstractions composed of a finite 'vocabulary' of low level control actions.
Inspired by such observations, we apply this cognitive neuroscience approach to the problem of sub-structure discovery in Hierarchical Reinforcement Learning (HRL) by making use of grammatical inference. More specifically, the HRL agent uses grammar induction to extract hierarchical constituents from trajectory 'sentences' (i.e. sequences of observed actions). Following this path we propose a solution to the credit assignment that is split into two alternating stages of inference (see left-hand side of figure \ref{fig:loop_ag}):

\begin{enumerate}
	\item \textbf{Grammar Learning}: Given episodic trajectories, we treat the time-series of transitions as a sentence sampled from the language of the policy-conditioned environment. Using grammar induction algorithms \cite{Manning_1997} the agent extracts hierarchical constituents of the current policy. Based on the estimated production rules, temporally-extended actions are constructed which convey goal-driven syntactic meaning. The grammar can efficiently be inferred (in linear time) and provides enhanced interpretability.
	\item \textbf{Action Learning}: Using the grammar-augmented action space, the agent acquires new value information by sampling reinforcement signals form the environment. They refine their action-value estimates using Semi-Markov Decision Process (SMDP) Q-Learning \cite{Bradtke_1995, Parr_1998a}. By operating at multiple time scales, the HRL agent is able to overcome difficulties in exploration and value information propagation. After action learning, the agent again samples simulated sentences by rolling out transitions from the improved policy.
\end{enumerate}

By alternating between stages of grammar and action value learning the agent iteratively reflects and improves on their habits in a semi-supervised manner. The inferred grammar parse trees are easy to interpret and provide semantically meaningful sub-policies.  
Our experiments highlight the effectiveness of the action grammars framework for imitation, curriculum and transfer learning given an expert policy rollout. Furthermore, we show promising results for an online implementation which iteratively refines grammar and value estimates. To the best of our knowledge, we are the first to introduce a general grammar-based framework that merges the fields of sequential decision making and grammatical inference. We provide a computationally efficient as well as interpretable account to the discovery of temporal abstractions. \\
The remainder of this report is structured as follows: First, we summarize and contrast the related literature. Next, we introduce the relevant technical background which includes learning with macro-actions in SMDPs as well as context-free grammar (CFG) inference. Afterwards, we introduce our proposed action grammars framework. We differentiate between two main paradigms: \textit{(a)} Learning an action grammar from expert policy rollouts and \textit{(b)} learning an action grammar from noisy on-policy rollouts acquired in an online fashion. Finally, we validate the proposed algorithmic framework on multiple challenging sparse credit assignment problems.   

\section{Related Work}

Hierarchical control of temporally-extended actions allows the RL agent to constrain the dimensionality of the temporal credit assignment problem. Instead of having to make an action choice at every tick of the environment, the top-level policy selects a lower-level policy that executes actions for potentially multiple time-steps. Once the lower-level policy finishes execution, it returns control back to top-level policy. Identification of suitable low level sub-policies poses a key challenge to HRL. 
The current state-of-the-art approaches can be grouped into three main pillars:
Graph theoretic \citep{Hengst_2002, Mannor_2004, Simsek_2004b} and visitation-based \citep{Stolle_2002, Simsek_2004b} approaches aim to identify bottlenecks within the state space. Bottlenecks are regions in the state space which characterize successful trajectories. This work, on the other hand, identifies patterns solely in the action space and does not rely on rewardless exploration of the state space. Furthermore, the proposed action grammar framework defines a set of macro-actions as opposed to full option-specific sub-policies. Thereby, it is less expressive but more sample-efficient to infer.
Gradient-based approaches, on the other hand, discover parametrized temporally-extended actions by iteratively optimizing an objective function such as the estimated expected value of the log likelihood with respect to the latent variables in a probabilistic setting \citep{Daniel_2016} or the expected cumulative reward in a policy gradient context \citep{Bacon_2017, Smith_2018}. Grammar induction, on the other hand, infers patterns without supervision solely based on a compression objective. The resulting parse tree provides an interpretable structure for the distilled skill set.
Futhermore, recent approaches \citep{Vezhnevets_2017, Florensa_2017} attempt to split the goal declaration and goal achievement across different stages and layers of the learned architecture. Usually, the top level of the hierarchy specifies goals in the environment while the lower levels have to achieve such. Again, such architectures lack sample efficiency and easy interpretation. The context-free grammar-based approach, on the other hand, is a symbolic method that requires few rollout traces and generalizes to more difficult task-settings. 

Finally, unlike recent work on unifying symbolic and connectionist methods, we do not aim to discover relationship between objects \citep{Garnelo_2016, Garnelo_2019, Zambaldi_2018}. Instead our proposed action grammar framework achieves interpretability by extracting hierarchical subroutines associated with sub-goal achievements.

\section{Technical Background}
Our method extends Deep Reinforcement Learning algorithms by the means of sub-skill distillation via grammatical inference. We briefly review the most simple form of temporally-extended actions, namely macro-actions, and context-free grammar inference which we fuse in the subsequent section.

\subsection{Temporal Abstractions and DQNs}
The HRL agent overcomes exploration problems, by restricting their decision making process in a syntactically meaningful way. Thereby, fewer decisions are evaluated. This notion of temporal abstraction can be formulated within the framework of SMDPs.
They extend Markov Decision Processes to account for not only reward and transition but also time uncertainty. The time between individual decisions is modeled as a random positive integer variable, $\tau \in \mathbb{Z}^{++}$. The waiting time is characterized by the joint likelihood of transitioning from state $s \in \mathcal{S}$ to state $s'$ in $\tau$ time steps given action $m$ was pursued, $P(s', \tau| s, m)$. Thereby, SMDPs allow one to elegantly model the execution of actions which extend over multiple time-steps. A macro-action \cite{McGovern_1997}, $m \in \mathcal{M}$ specifies the sequential and deterministic execution of multiple ($\tau_m$) primitive actions. Primitive actions can trivially be viewed as one-step macro-actions. Let $r_{\tau_m} = \sum_{i=1}^{\tau_m} \gamma^{i-1} r_{t+i}$ denote the accumulated and discounted reward for executing a macro. Tabular value estimates can then be updated using SMDP-Q-Learning \cite{Bradtke_1995, Parr_1998a} in a model-free bootstrapping-based manner:
\begin{equation}
	Q(s, m)_{k+1} = (1-\alpha) Q(s, m)_k + \alpha \left( r_{\tau_m} + \gamma^{\tau_m} \max_{m' \in \mathcal{M}} Q(s', m')_k \right)
\end{equation}

In order to increase sample efficiency one can perform intra-macro updates for each state transition tuple $\{<s,a,r,s'>\}_{\tau_m}$ within the macro execution. Without restricting validity we can generalise all these results to Deep RL, for example, in the DQN \citep{Mnih_2015} framework. Then, the DQN Bellman error objective can be adapted to the semi-Markov case:

\begin{equation}
\mathcal{L}(\theta) := \mathbb{E} [(r_{\tau_m} + \gamma^{\tau_m} \max_{m' \in \mathcal{A} \cup \mathcal{M}} Q(s',m';\theta^-) - Q(s,m; \theta))^2] \end{equation}

The gradient with respect to the parameters is approximated by samples from the experience replay (\cite{Lin_1992}; ER) buffer, $\{s,m,r_{\tau_m},s', \tau_m\} \sim D_{\mathcal{M}}$. The learning dynamics are stabilized by using a separate target network \citep{Mnih_2015}. Furthermore, all recent extensions to DQNs apply (see e.g. \citep{Hessel_2018}).

\subsection{Context-Free Grammar Inference} 
Just like language-based communication between sender and receiver, action sequences convey goal-directed semantic meaning relating task, environment and actor. Action sequences reflect common hierarchical structures induced by task goal, agent and environmental constraints. Many real world problems require a hierarchy of sub-goal achievements which increase in sequential complexity and timescale duration. Therefore, we need to define our grammar-based approach formally: Formal grammars (formal language theory \citep{Chomsky_1959a}) study generating (grammar) and accepting systems (automatons) that underlie a language. Given a start symbol $S$, a formal grammar $(\Sigma, \mathbb{N}, S, \mathcal{P})$ derives an output of strings. Production rules $\mathcal{P}$ map a set of non-terminal vocabulary $\mathbb{N}$ either to another non-terminal or terminal string within the terminal vocabulary $\Sigma$.
Context-free grammars (CFG) \citep{Chomsky_1959a} constrain the set of productions to either map from one-to-one, one-to-none or one-to-many. A non-branching and loop-free CFG is called a straight-line grammar.

Ultimately, the HRL agent shall learn action grammars from observing patterns in its action sequences. The process of inferring a grammar for a language that is consistent with a given sample of sentences is termed grammar induction \citep{Levelt_2008}. We use two different approaches to highlight that our method is robust with respect to the underlying grammar learner.
A simple solution is the Sequitur algorithm \citep{Manning_1997}: Given a single sentence of the language, it sequentially reads in all symbols and collects repeating sub-sequences of symbols into a production rule. Therewhile, the final encoded string is only allowed to have unique bigrams (\textit{Digram Uniqueness}, \citep{Manning_1997}) and production rules must be used more than once in the derivation of the string (\textit{Rule Uniqueness}, \citep{Manning_1997}).
In order to overcome Sequitur's problem of noise overfitting, $k$-Sequitur \citep{Stout_2018} has been proposed. Instead of replacing a bigram with a rule if the bigram occurs twice, it has to occur at least $k$ times. As $k$ increases the discovered CFG grammar becomes less sensitive to overfitting noise and the resulting grammar is more parsimonious in terms of productions. 
An alternative approach, Lexis \cite{Siyari_2016b}, provides an optimization-based procedure which iteratively constructs a directed acyclic graph (DAG). Starting from a trivial graph which connects a set of target sentences with the set of elements in the terminal vocabulary, the Lexis-DAG is constructed by adding intermediate nodes. This problem by itself is NP-hard. G-Lexis, the greedy algorithmic implementation, searches for sub-strings that will lead to a maximal reduction in the cost, when added as new intermediate node.

\section{Context-Free Action Grammars}
Having laid the foundations of both temporal abstractions in RL and grammatical inference and, we state our fundamental assumption that connects formal language theory with goal-oriented  sequential behavior as follows:

\begin{assumption}
	Observed episodic behavior (with trajectory $\vartheta = \{\vartheta_1, \dots, \vartheta_T\}$ where $\vartheta_t =\{s_t, a_t\}$) can be equivalently viewed as sentences sampled from the language, $L(G)$ with $G \sim \pi|E$.
\end{assumption}

A trajectory obtained from traversing the current policy $\pi_k$ within the environment $E$ can be viewed as a sample from the \emph{policy language} generated by the policy-specific grammar, $L(\pi_k|E)$.
Let the terminal vocabulary $\Sigma$ consist of the primitive action space $\mathcal{A}$, $\Sigma = \mathcal{A}$. We denote  $\vartheta^i \sim L(\pi_k|E)$ for $i = 1, \dots N_g$ trajectories. Given a set of trajectories, a CFG estimate $\hat{G}$ can be inferred and the resulting production rules transformed into macro-actions $\mathcal{M}^{\hat{G}}$ by recursively flattening the non-terminals. We propose to augment the action space of the agent such that $\mathcal{A}^{\hat{G}} = \mathcal{A} \cup \mathcal{M}^{\hat{G}}$. 
\begin{figure}[!b]
\centering
    \includegraphics[width=0.55\linewidth]{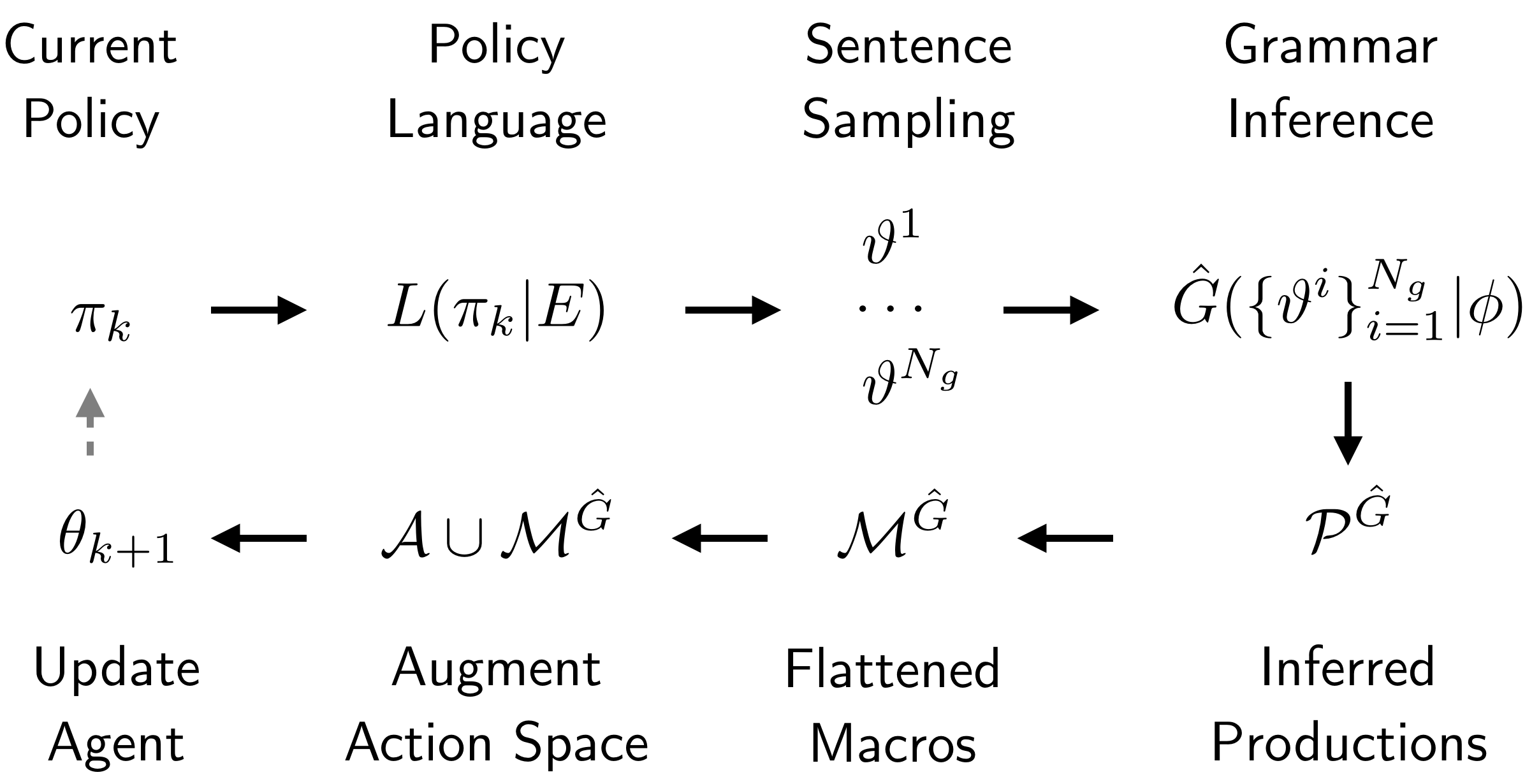}
    \caption{A closed-loop for alternating between on-policy rollouts, grammar induction and value-based learning. Given a policy $\pi_k$ the agent generates on-policy rollouts $\vartheta_i$. Based on such sampled sentences, the agents infers a context-free grammar $\hat{G}$. By flattening the resulting production rules, we obtain a set of macro-actions $\mathcal{M}^{\hat{G}}$. After augmenting the action spaces of the HRL agent with the inferred temporal abstractions, the agent continues to refine their value estimates (parametrized by $\theta_k$). Given the context of the credit assignment problem, the agent may iterate the grammar inference procedure with a new 'greedified' policy (online inferred grammar). For the expert and transfer learning RL case only one grammar (based on $\pi_\star^\mathcal{T}$ or $\pi_\star^{\mathcal{T}'}$) is inferred at the learning onset.}%
  \label{fig:ag_loop}
\end{figure}
Afterwards, the HRL agent refines their action-value estimates. A schematic illustration of this procedure is shown in figure \ref{fig:ag_loop}. Depending on the generating policy of the compressed traces, we propose different grammar-based HRL agents (see right-hand side of figure \ref{fig:loop_ag}).

\subsection{Action Grammars for Imitation and Transfer Learning} 

If the traces $\vartheta^i$ are sampled from the language $L(\pi_\star^{\mathcal{T}}|E)$ generated by the optimal policy $\pi_\star^{\mathcal{T}}$, the agent can use the resulting grammar macros in an imitation learning \citep{Bain_1999} setting. Before the onset of the first value learning stage, the action space is augmented with the optimal flattened productions. Thereby, the task-specific knowledge is distilled and transferred from a teacher policy to the student.
Furthermore, an agent faced with learning a curriculum of tasks can make use of the optimal grammar of an easier solved task, $\mathcal{T}'$. Skills universal to all tasks do not have to be re-learned at every stage. Instead, the inferred optimal grammar provides an effective knowledge structure which accelerates the agents learning process.

An intuitive illustration of a HRL problem which requires the repeated usage of skills is given by the general Towers of Hanoi (ToH) problem (see figure \ref{fig:hanoi}). In order to solve the $N$-disk ToH problem the agent has to identify a hierarchical and recursive principle. By moving $n-1$ disks onto an auxiliary pole and the $n$-th disk onto the target pole, the agent is able solve the sparse reward problem. Since such a routine can easily be formulated within a grammar parse tree, we hypothesize that the action grammars framework may provide an efficient solution.
The problem is formulated as a sparse long-term credit assignment problematic. The size of the state space grows exponentially, $|S| = 3^N$ (all possible allowed orderings), and the optimal number of moves to solve the game is given by $2^N - 1$. 

 \begin{minipage}{\textwidth}
  \begin{minipage}[b]{0.49\textwidth}
    \centering
    \includegraphics[width=\textwidth]{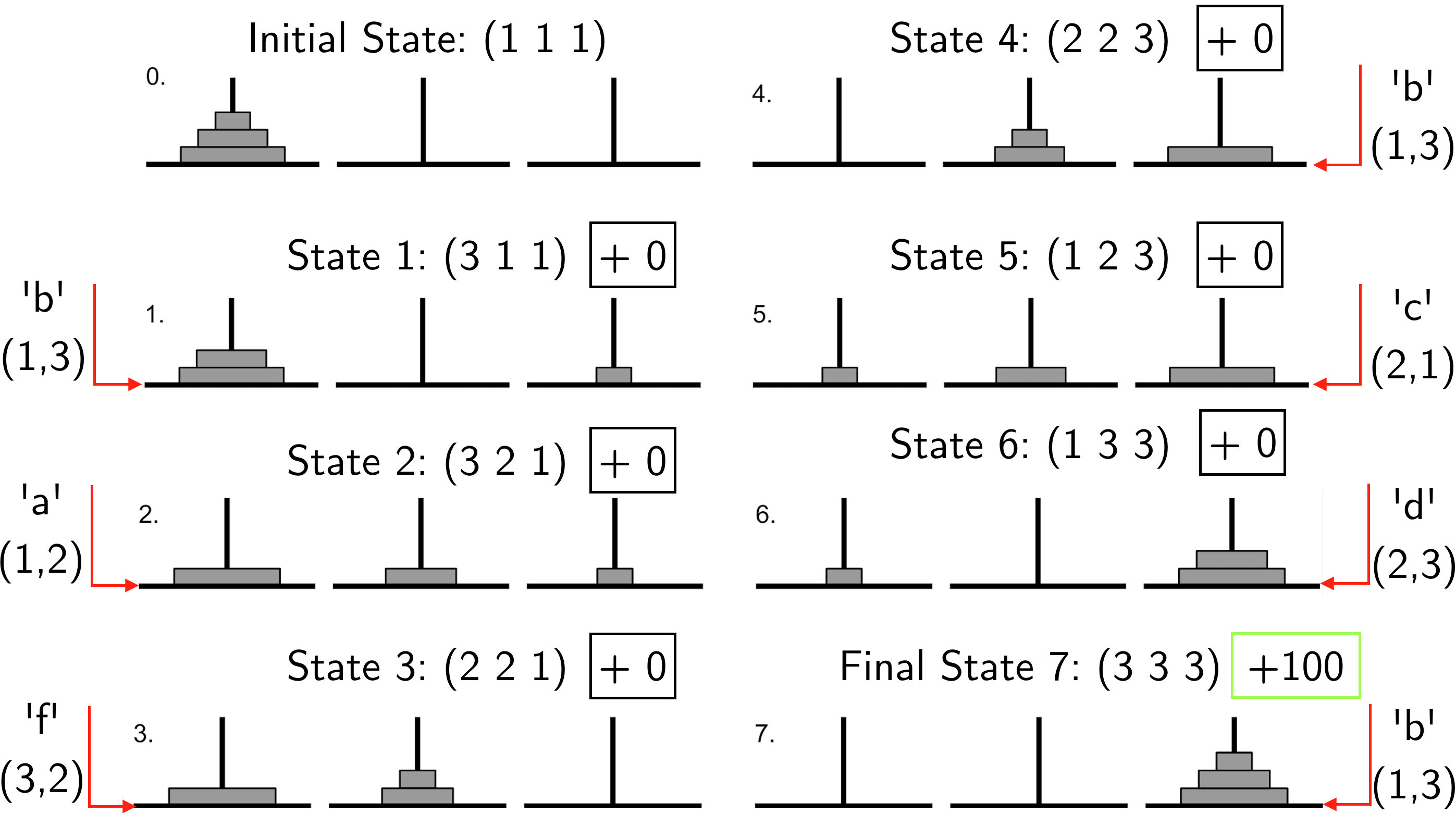}
    \captionof{figure}{RL Formulation of the ToH Problem.}
    \label{fig:hanoi}
  \end{minipage}
  \rulesep
  \begin{minipage}[b]{0.49\textwidth}
    \centering
    \begin{tabular}{c | c c | c c } 
\hline\hline 
 \multicolumn{5}{c}{Trace ($\vartheta^\star$): bafbcdbafecfbafbcdbcfecdbafbcdb
} \\
\hline\hline
  & \multicolumn{2}{c}{$2$-Sequitur} & \multicolumn{2}{c}{G-Lexis}\\
\hline
 $\vartheta^{enc}$ & \multicolumn{2}{c}{BCDfBEfDdBb} & \multicolumn{2}{c}{BbafecfBbcfecdBb}\\
 $\frac{|\vartheta|}{|\vartheta^{enc})|}$ & \multicolumn{2}{c}{2.817} & \multicolumn{2}{c}{1.938}\\
\hline \hline
$\mathbb{N}$ & $\mathcal{P}^{2-S}$ & $\mathcal{M}^{2-S}$ & $\mathcal{P}^{G-L}$ & $\mathcal{M}^{G-L}$\\ 
\hline 
B & CEd & bafbcd & bafbcd & bafbcd\\
C & baf & baf &&  \\
D & ec & ec &&  \\
E & bc & bc &&  \\ 
\hline 
\end{tabular}
      \captionof{table}{ToH (5 disks) Grammar-Macro Construction.}
      \label{table:optimal_grammar}
    \end{minipage}
  \end{minipage}
  
Each of the possible primitive actions can be encoded as a symbol resulting in the following terminal vocabulary: $\Sigma = \mathcal{A} = \{a, b, c, d, e, f\}$. Table \ref{table:optimal_grammar} depicts the grammars and resulting macros inferred from a trace of the optimal policy 5-disk ToH problem using the 2-Sequitur and G-Lexis algorithm. 
The optimal policy trace $\vartheta$ is compressed into $\vartheta^{enc}$. The respective production rules $\mathcal{P}$ are flattened (if possible) to obtain the macro-set $\mathcal{M}$. We can observe that $2$-Sequitur encodes more aggressively, yielding more and shorter production rules compared to G-Lexis.  
The 2-Sequitur expert grammar HRL agent's action space is augmented as follows:

\begin{equation}
\mathcal{A}^{\hat{G}} = \mathcal{A} \cup \mathcal{M}^{2-S} = \{a,b,c,d,e,f\} \cup \{bafbcd, baf, ec, bc\}
\end{equation}

The augmented action space can then be used to refine the macro-action value estimates with the help of tabular SMDP-Q-Learning (see equation 1). The same general paradigm translates into the extended value function approximator setting. Most importantly the knowledge endorsement is performed by effectively enhancing the action space of the agent and not by transferring function approximation parameters. Hence, we do not require structural similarity in the representation of task-specific observations.

\vspace{-0.25cm}
\subsection{Action Grammars for Online Learning} 

\begin{figure}[b!]
\minipage{0.32\linewidth}
  \includegraphics[width=\linewidth]{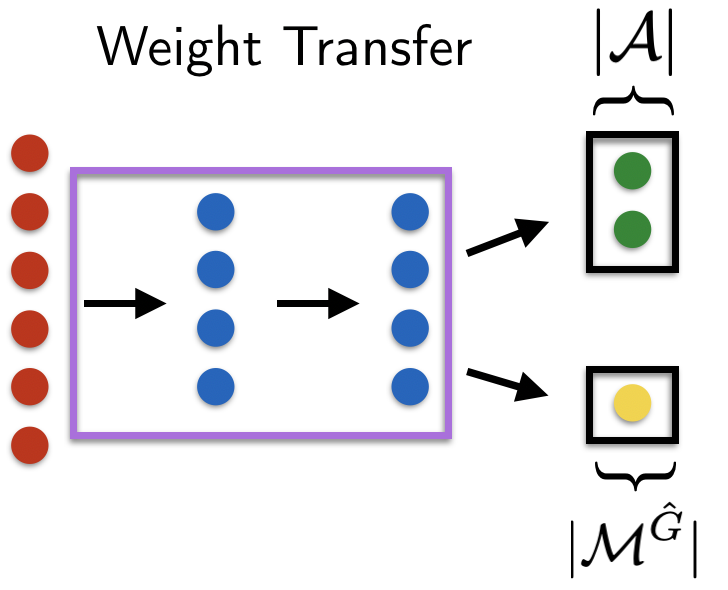}
\endminipage
\rulesep
\minipage{0.32\linewidth}
  \includegraphics[width=\linewidth]{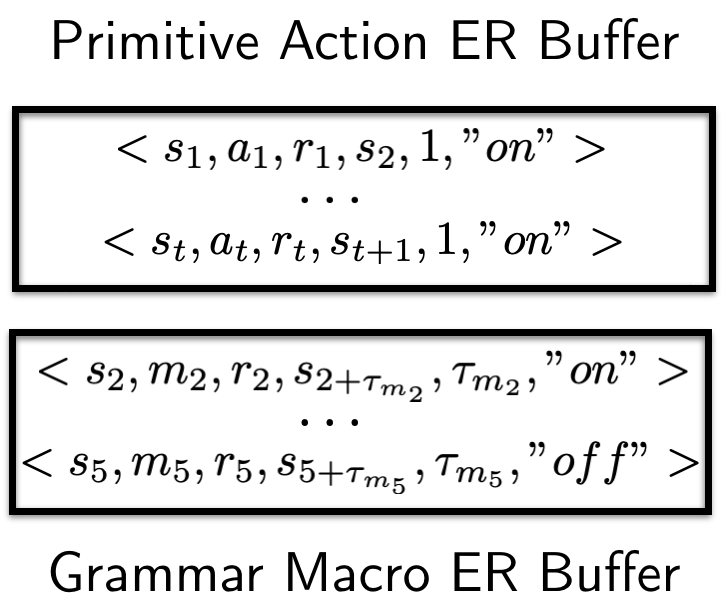}
\endminipage
\rulesep
\minipage{0.32\linewidth}
  \includegraphics[width=\linewidth]{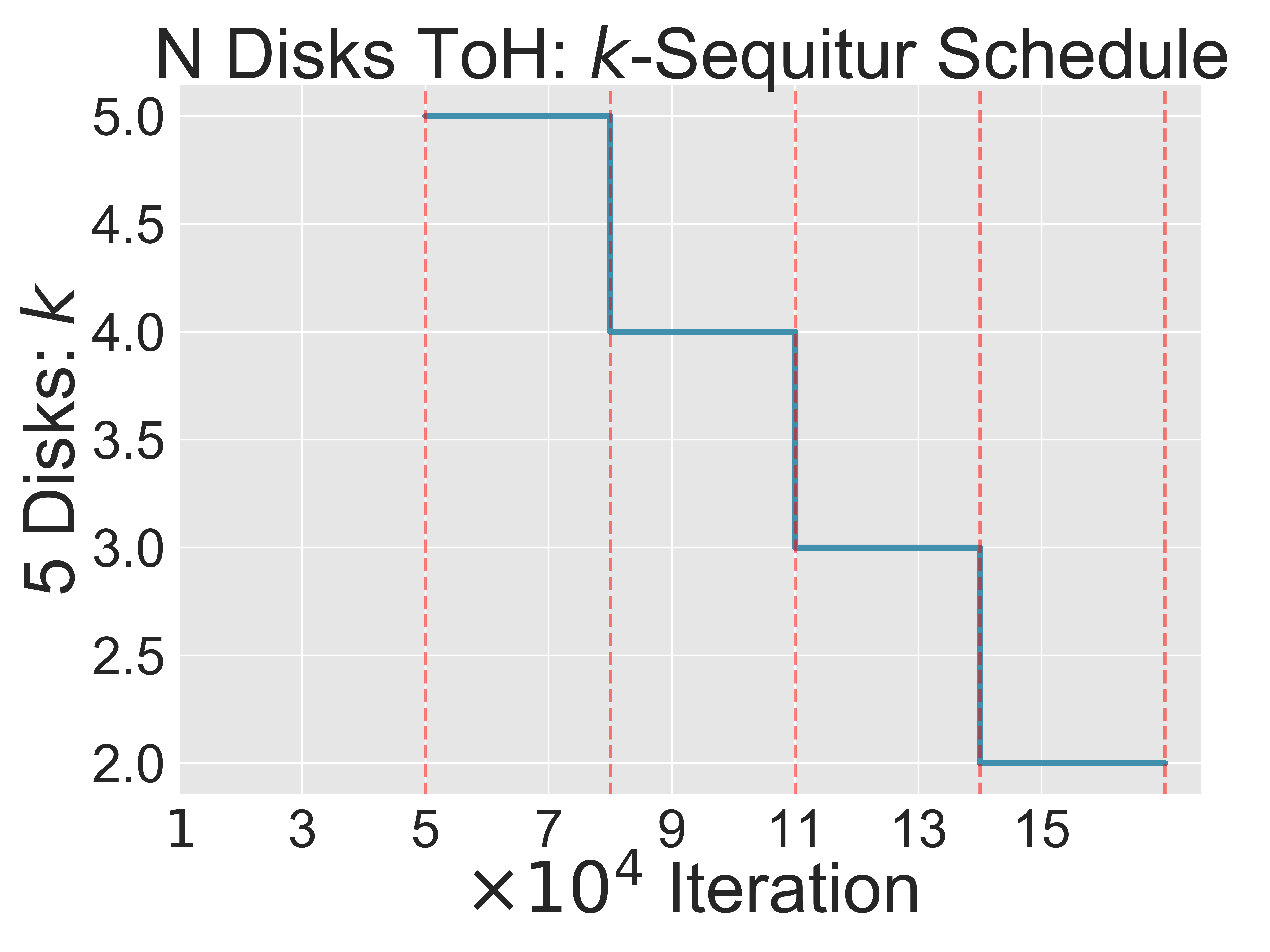}
\endminipage
\caption{Online Action Grammar: \textbf{Left.} Grammar Transfer DQN. \textbf{Middle.} Grammar Replay Buffer. \textbf{Right.} Grammar Inference Hyperparameter Schedule ($\phi=k$). Red vertical lines indicate a grammar update.}
\label{fig:online_adaptations}
\end{figure}

In the previous section we have introduced an initial grammar induction step to distill semantically meaningful sub-policies from optimal behavior. Here we extend the formalism to online \textit{grammar self-induction}.
If an episode successfully terminates, the grammar inference process identifies repeating sub-goal achieving patterns. We hypothesize that by extracting action grammar sub-sequences, one compresses the temporal dimension of the credit assignment problem and identifies successful habits.
After each grammar compression step, the action space is augmented by the set of grammar macros. 
The set of previously inferred macros becomes inactive. In order to preserve value estimates between updates and increase sample efficiency, we propose a set of practical solutions:

(1) \textit{Transfer learning} \citep{Oquab_2014}: To accommodate the variable set of grammar-inferred skills, the size of the DQN output layer has to be updated. Transferring the value-relevant feature detector parameters between action space augmentation, allows the agent to use the previously learned value characteristics (see left-hand side of figure \ref{fig:online_adaptations}).

(2) \textit{Grammar Experience Replay Buffer}: The ER replay buffer will contain experiences associated with currently "inactive" grammar macro-actions. Therefore, it is necessary to maintain a grammar-altered buffer system, where we store a separate indicator ("on"/"off") for whether a macro-action is currently part of the active grammar-augmented action space. By default all primitive actions are set to "on".
At any given point the agent can only sample macro transitions which are associated with the currently active set (see middle part of figure \ref{fig:online_adaptations}). Thereby, sample efficiency is increased once a grammar macro is repeatedly inferred.

(3) \textit{Intra-Macro Updates}: During the execution of a macro-action, we store the overall macro transition tuple $<s_t,m_t,r_{t +\tau_m}, s_{t +\tau_m +1}, \tau_m, "on">$ as well as the individual transitions $\{<s_i,a_i,r_i, s_{i+1}, 1, "on">\}_{i=t}^{t + \tau_m}$. Hence, we replay primitive actions proportionately more frequently. This has the advantage of increasing the robustness of the learned policy to noise in the transitions. The agent replays all gathered one-step transition experiences throughout the overall learning process.

(4) \textit{Grammar Induction Hyperparameter Schedule}: The length of the sampled action-sentences is directly linked to the progress an agent makes within the environment. 
If the goal of the agent is to quickly reach a certain state, the length will decrease over time. Hence, we might initially infer too many productions, unnecessarily blow up the action space and thereby diminish learning progress.
One possibility to overcome this problem is to only augment the action space by a set of top-$l$ most used flattened productions when encoding the final sequence (hence $|M^{\hat{G}}| = l$). 
Alternatively, we may control the parsimony by the means of the grammar hyperparameters $\phi$. In our experiments we experimented with a linear-decaying schedule for the regularization parameter $k$ of the $k$-Sequitur grammar inference algorithm (see right-hand side of figure \ref{fig:online_adaptations}). 

\vspace{-0.25cm}
\section{Experiments}
To demonstrate our action grammar RL approach we develop a series of experiments. The goal of the following experiments is to evaluate the following 3 key questions: 
(1) Does a grammar learned from optimal policy rollouts allow for rapid imitation learning? (2)  Can CFGs distill and transfer task knowledge within a curriculum learning setting? (3)  Is online grammar inference able to structure the exploration process of the HRL agent? In order to answer these question we choose the general $N$-disk Towers of Hanoi environment (see figure \ref{fig:hanoi}) as well as a hierarchically structured gridworld task (see figure \ref{fig:gridworld}). We provide further experiment specifications in the supplementary material.

\subsection{Learning with Expert \& Transfer Action Grammars}

\begin{figure}[b!]
\minipage{0.32\linewidth}
  \includegraphics[width=\linewidth]{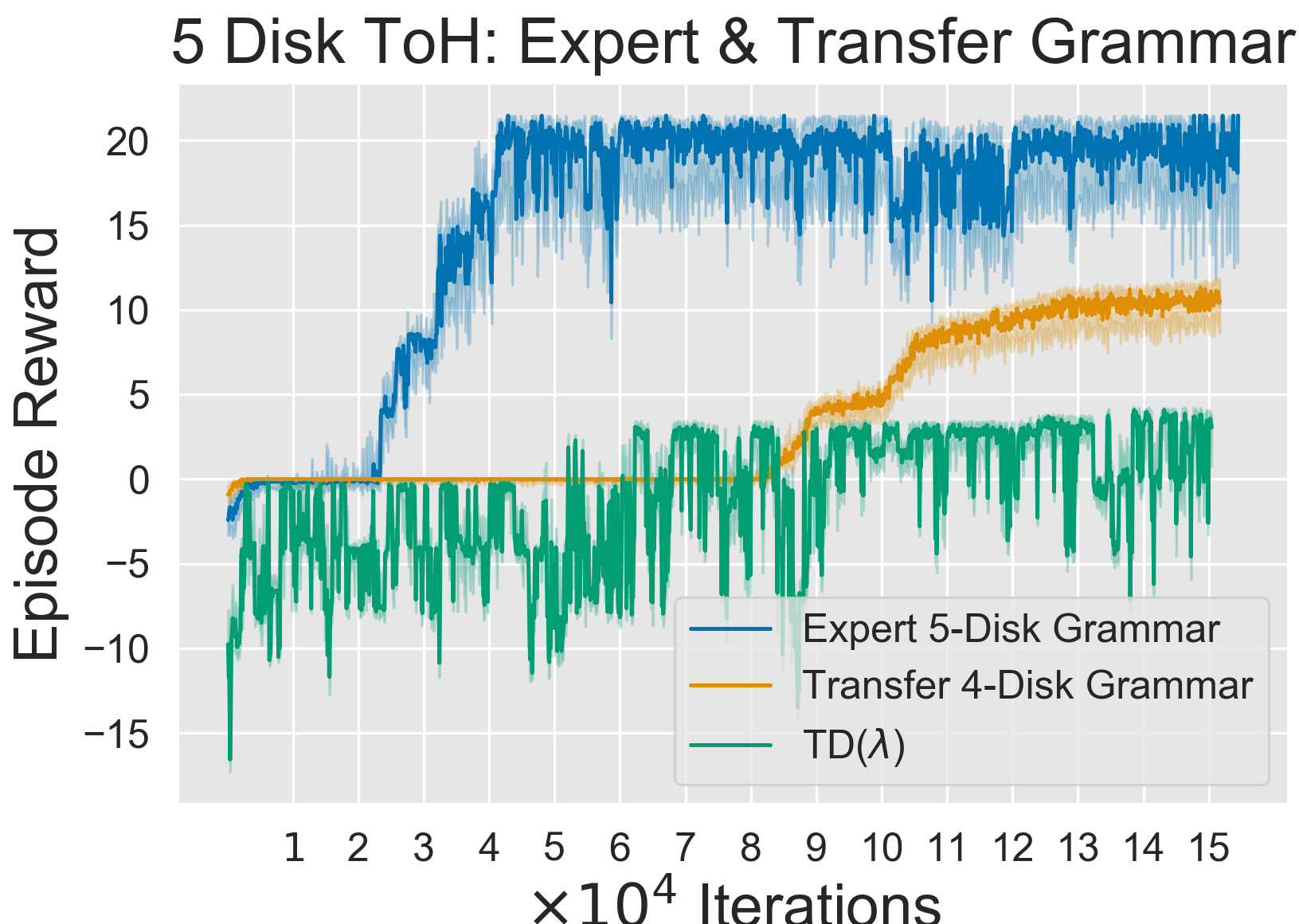}
\endminipage\hfill
\minipage{0.31\linewidth}
  \includegraphics[width=\linewidth]{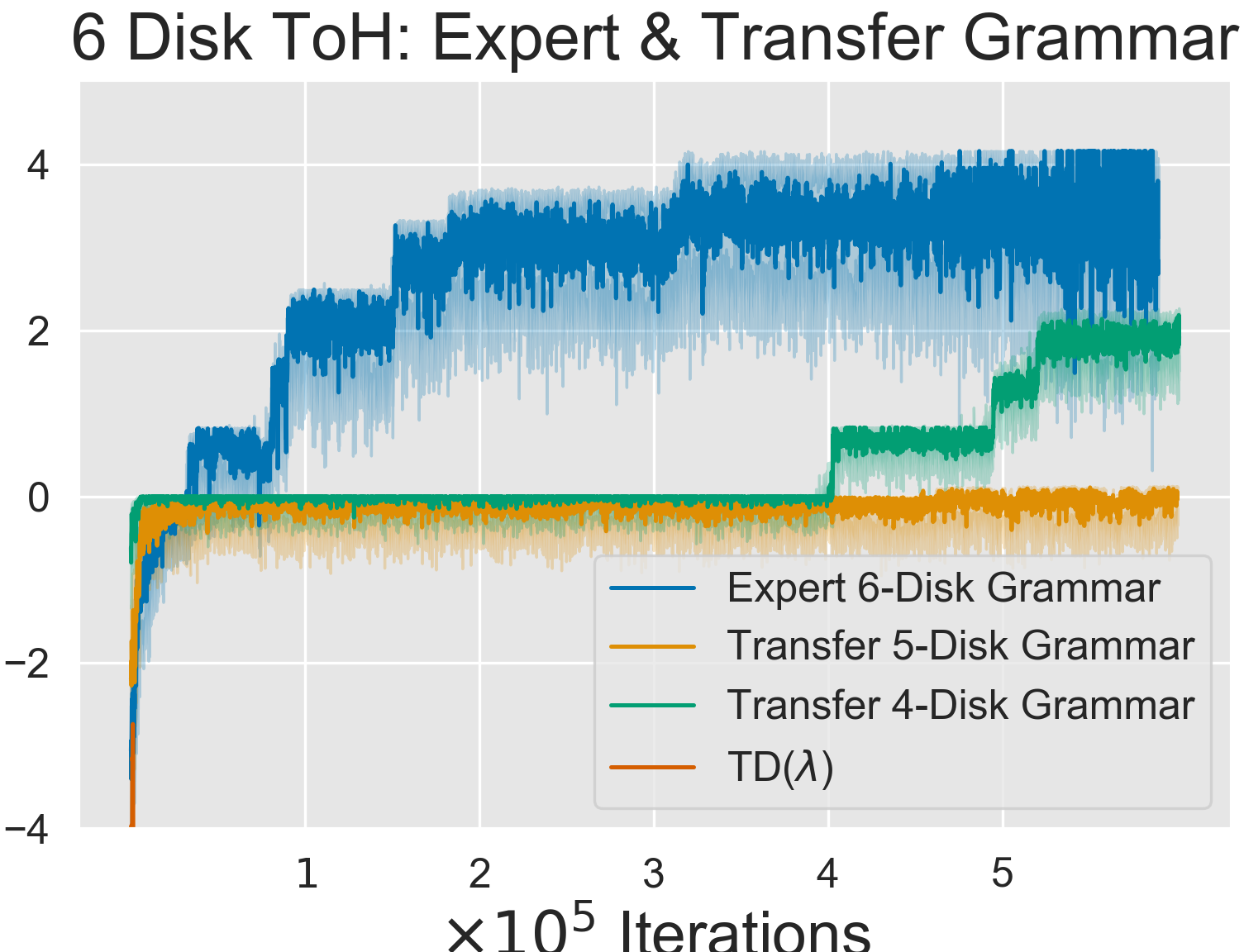}
\endminipage\hfill
\minipage{0.31\linewidth}
  \includegraphics[width=\linewidth]{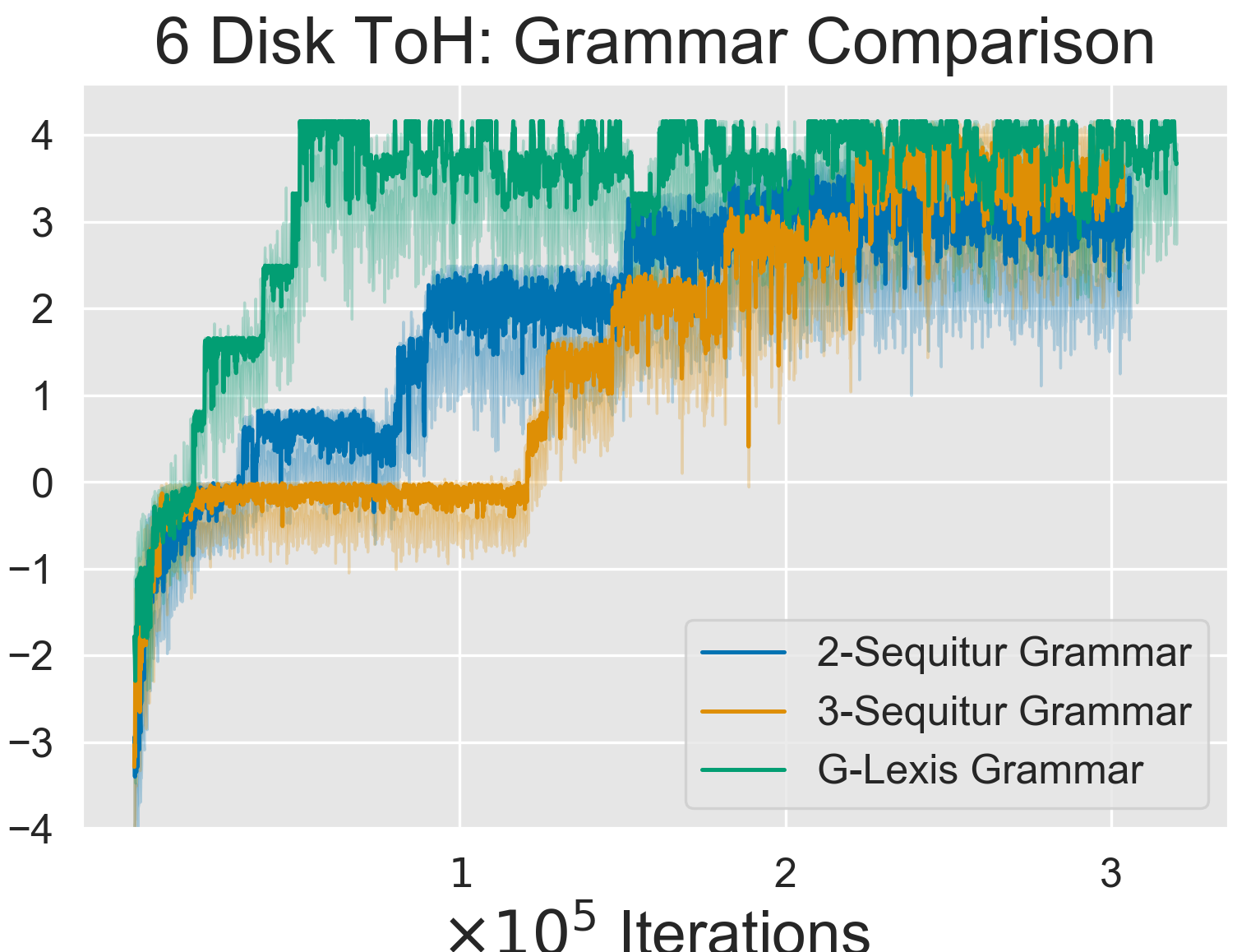}
\endminipage\hfill
\caption{Expert \& Transfer Grammar (\textit{ToH}): \textbf{Left.} 5 Disk Environment. \textbf{Middle.} 6 Disk Environment. \textbf{Right.} Comparison of $k$-Sequitur \& G-Lexis performance. Avg. over 5 random seeds. Median, 10th and 90th percentile.}
\label{fig:expert_grammar_toh}
\end{figure}

Figure \ref{fig:expert_grammar_toh} displays results for the $N$-disk ToH problem with different SMDP-Q-Learning agents and macro-actions defined by the production rules inferred from a single trace of the optimal policy.
The expert grammar macros significantly accelerate the learning progress and reduce the variance of policy rollouts. This provides evidence for our hypothesis that context-free grammars provide an effective temporal compression of the sequential credit assignment problem.
Furthermore, the transfer grammar agent is capable of transferring the knowledge distilled in a simpler optimal grammar (4 disks) to a more complex setting (6 disks). Finally, we compare different grammatical inference algorithms. We find that G-Lexis provides the most effective hierarchical compression of the 6 disk ToH problem solution. When inspecting the corresponding sets of grammar macro-actions it becomes apparent that G-Lexis infers longer and fewer temporally-extended actions. 
\begin{figure}[b!]
\minipage{0.48\linewidth}
\centering
  \includegraphics[width=0.7\linewidth]{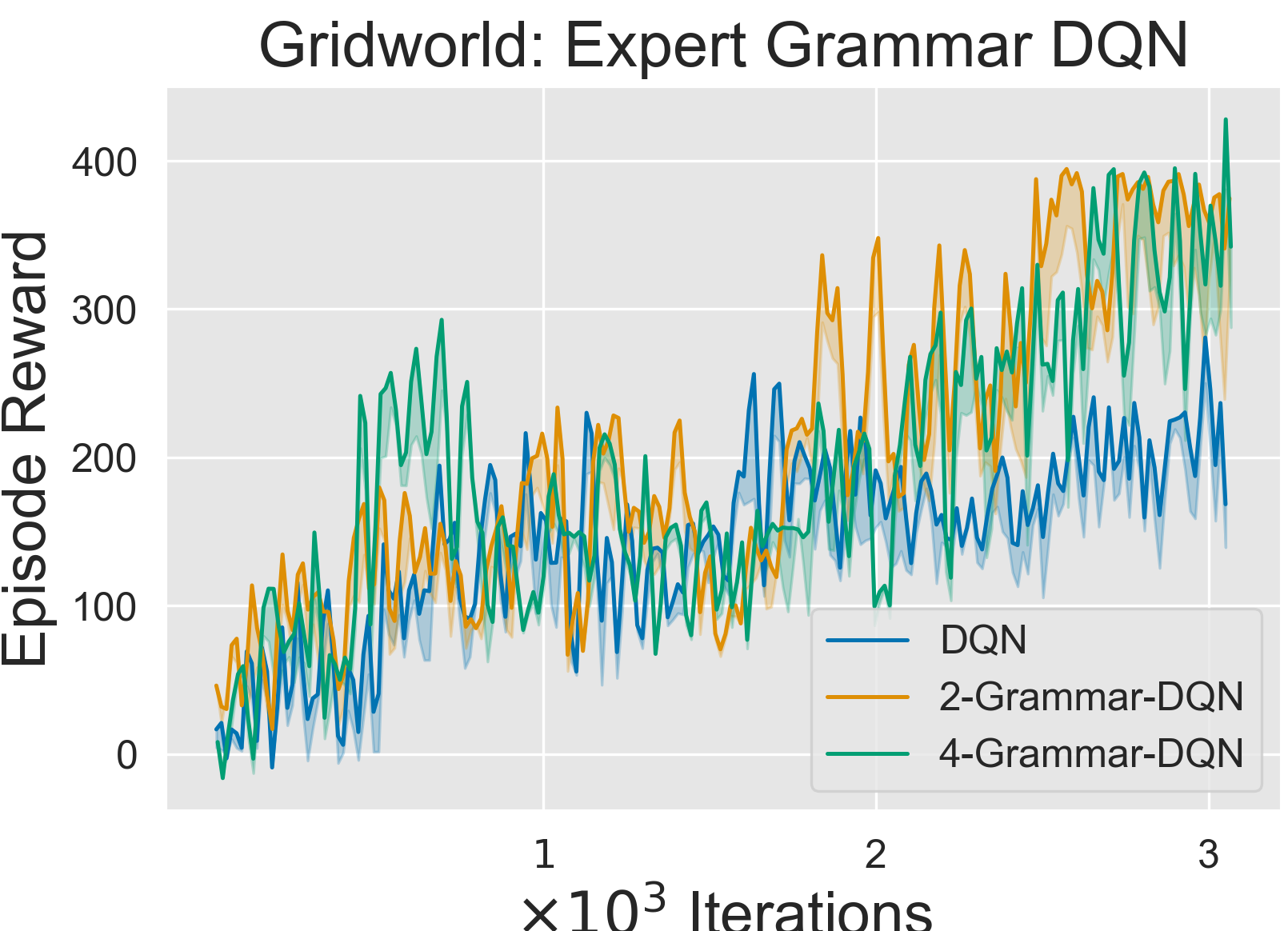}
\endminipage\hfill
\minipage{0.48\linewidth}
\centering
  \includegraphics[width=0.7\linewidth]{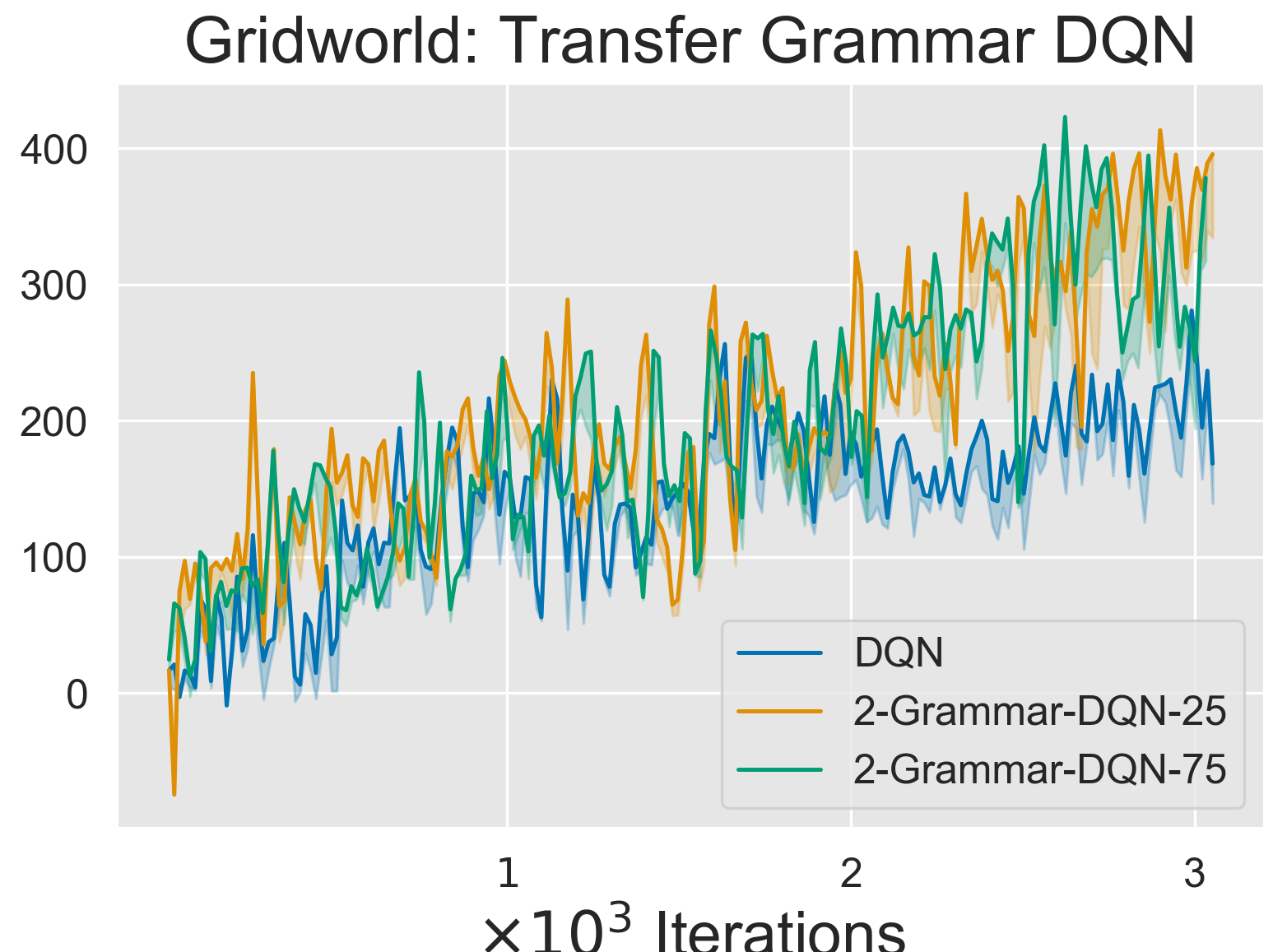}
\endminipage\hfill
\minipage{0.48\linewidth}
\centering
  \includegraphics[width=0.7\linewidth]{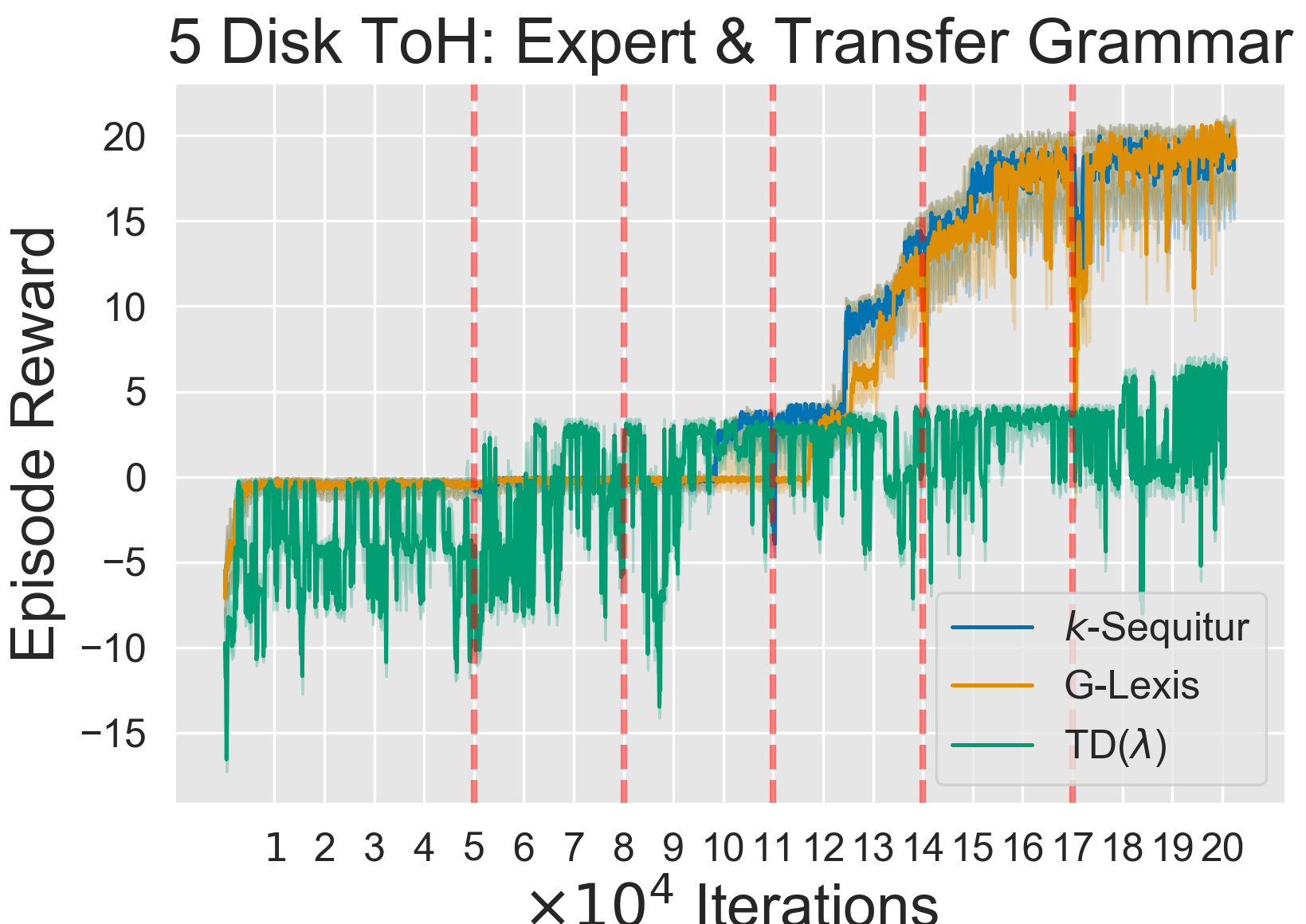}
\endminipage\hfill
\minipage{0.48\linewidth}
\centering
  \includegraphics[width=0.7\linewidth]{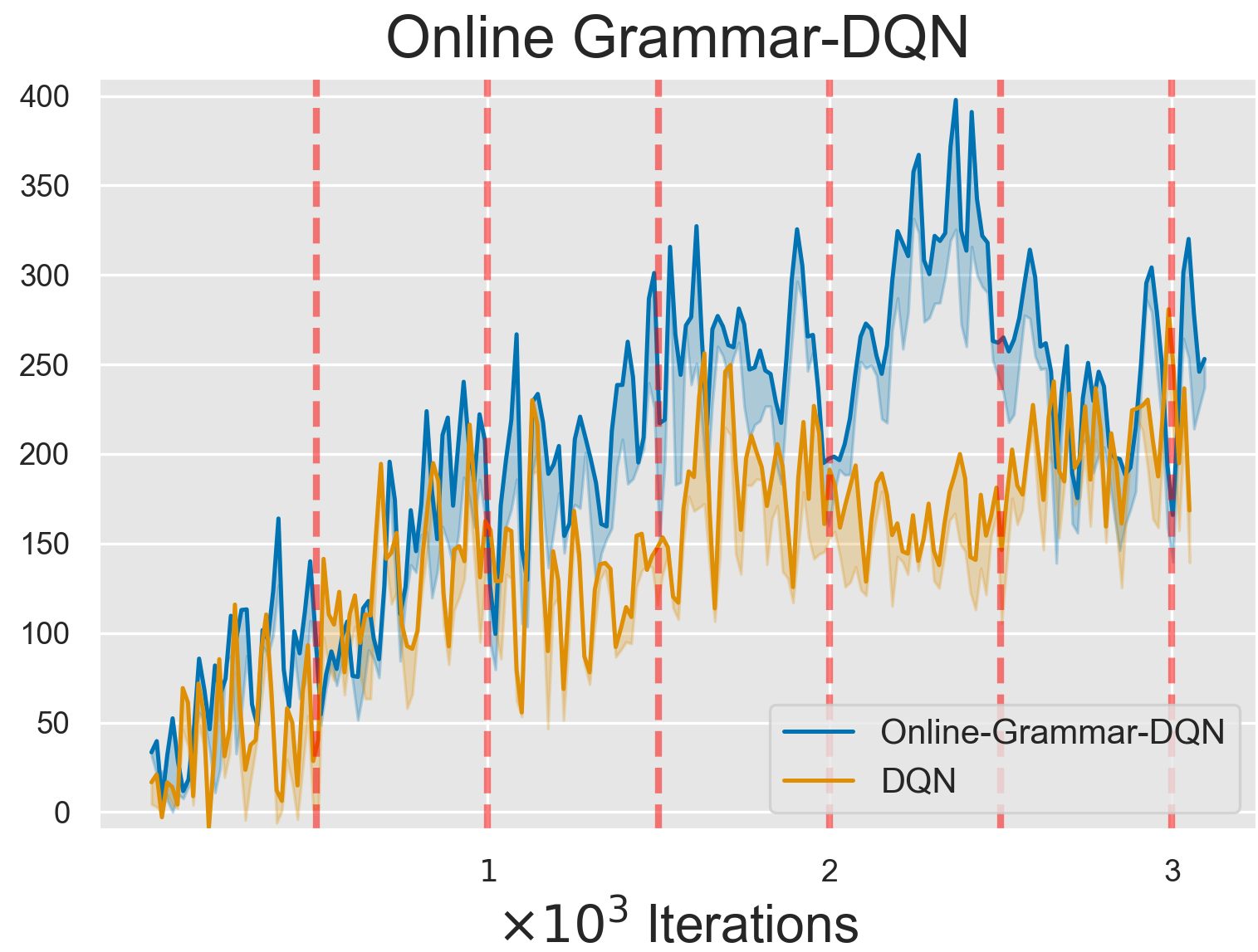}
\endminipage\hfill
\caption{\textbf{Top.} Expert \& Transfer Grammar (\textit{Gridworld}): \textbf{Left.} Expert Grammar (2 and 4 macros extracted from 2-Sequitur grammar inference). \textbf{Right.} Transfer Grammar. Averaged over 5 random seeds. Median, 10th and 90th percentile (2 macros extracted from DQNs trained for 25 and 75 episodes). \textbf{Bottom.} Online Inferred Grammars: \textbf{Left.} Visualization of the Grammar Macro "B" for the 5 disk ToH problem. \textbf{Right.} Grammar Inference Compression and Convergence Statistics. Horizontal lines correspond to optimal (policy rollout) compression and entropy.}
\label{fig:expert_grammar_grid}
\end{figure}
The gridworld Grammar-DQN agent (see top row of figure \ref{fig:expert_grammar_grid}) again infers a set of macro-actions from a single expert rollout. Afterwards, the output layer and action space are augmented. 
The two Expert Grammar-DQN agents differ in the amount of macro-actions (top two and four most used productions in the encoded policy trace) inferred with 2-Sequitur on a converged DQN agent rollout. Again, the expert grammar-endorsed agent is accelerated in their learning progress.
The two Transfer Grammar-DQN agents, on the other hand, infer a set of two grammar macros from a single non-optimal DQN agent's (trained for 25 or 75 episodes) policy rollout. Our experiments show, that even with noisy non-optimal rollouts the inferred grammar allows the agent to exploit the inferred structure of the environment.

\subsection{Learning with Online Inferred Action Grammars}

The bottom row of figure \ref{fig:expert_grammar_grid} displays the results of the online grammar inference framework for both the ToH as well as the gridworld task. 
We infer a set of macro-actions after an initial warm-up phase (5000 Q-Learning iterations for ToH and 500 SGD updates for the Gridworld DQN).
The $k$ hyperparameter used to infer a set of grammars in ToH problem is given by the linear schedule in figure \ref{fig:online_adaptations}. For the Gridworld DQN agent we infer a new set of grammar macros from a self-rollout using 2-Sequitur throughout the entire learning process. 
We augment the action space with the top five (ToH) and two (Gridworld) most utilised flattened production rules in the trace compression. The learning dynamics provide a competitive extension to the general DQN framework and significantly outperform a $TD(\lambda)$ baseline.
We want to emphasize the relationship between grammar inference and exploration. In our experiments we found that the frequency of grammar updating as well as the grammar inference hyperparameters play an important role.

\subsection{Interpretability \& Convergence of the Inferred Action Grammars}

\begin{figure}[b!]
\minipage{0.49\linewidth}
  \includegraphics[width=\textwidth]{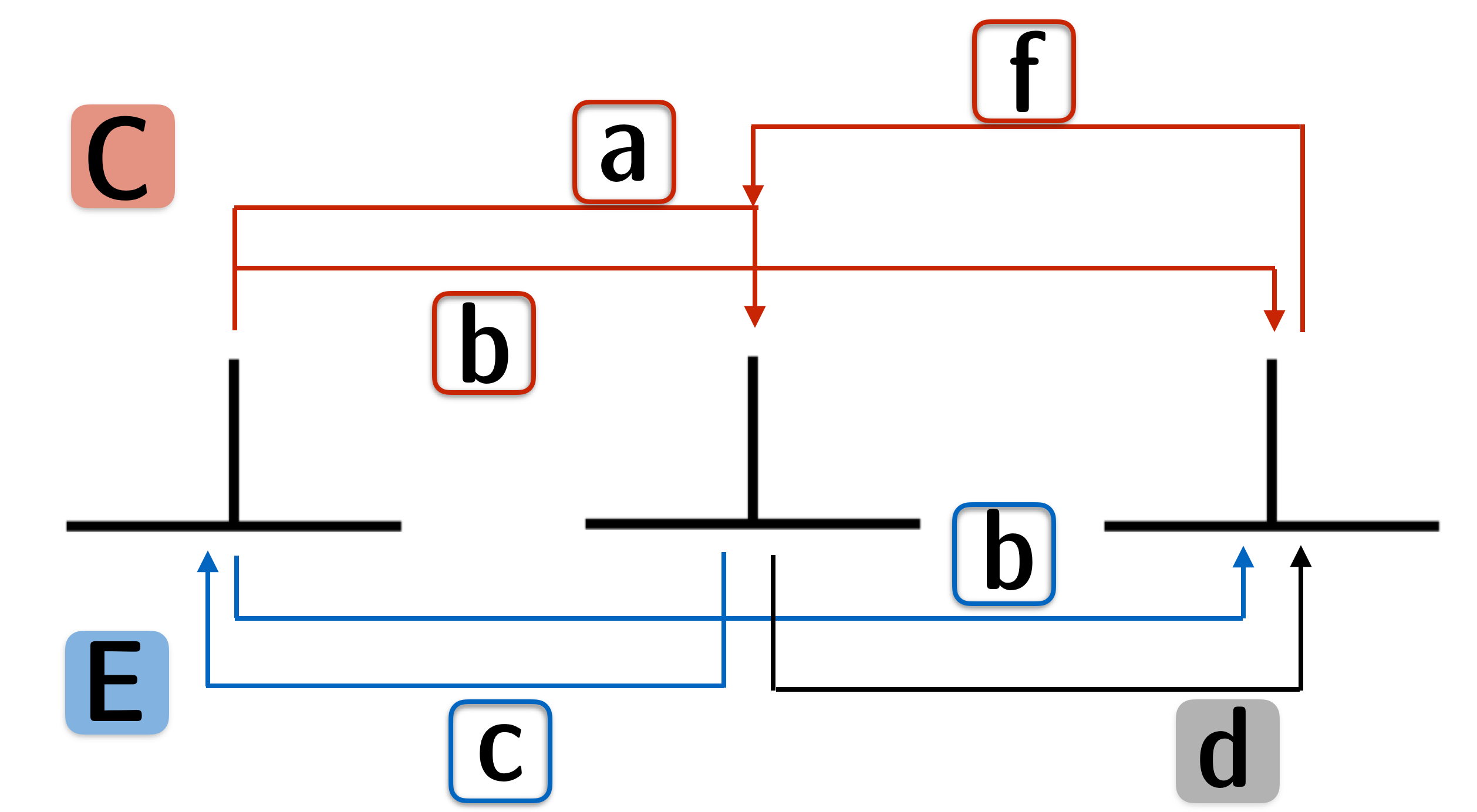}
\endminipage\hfill
\minipage{0.49\linewidth}
 \includegraphics[width=\textwidth]{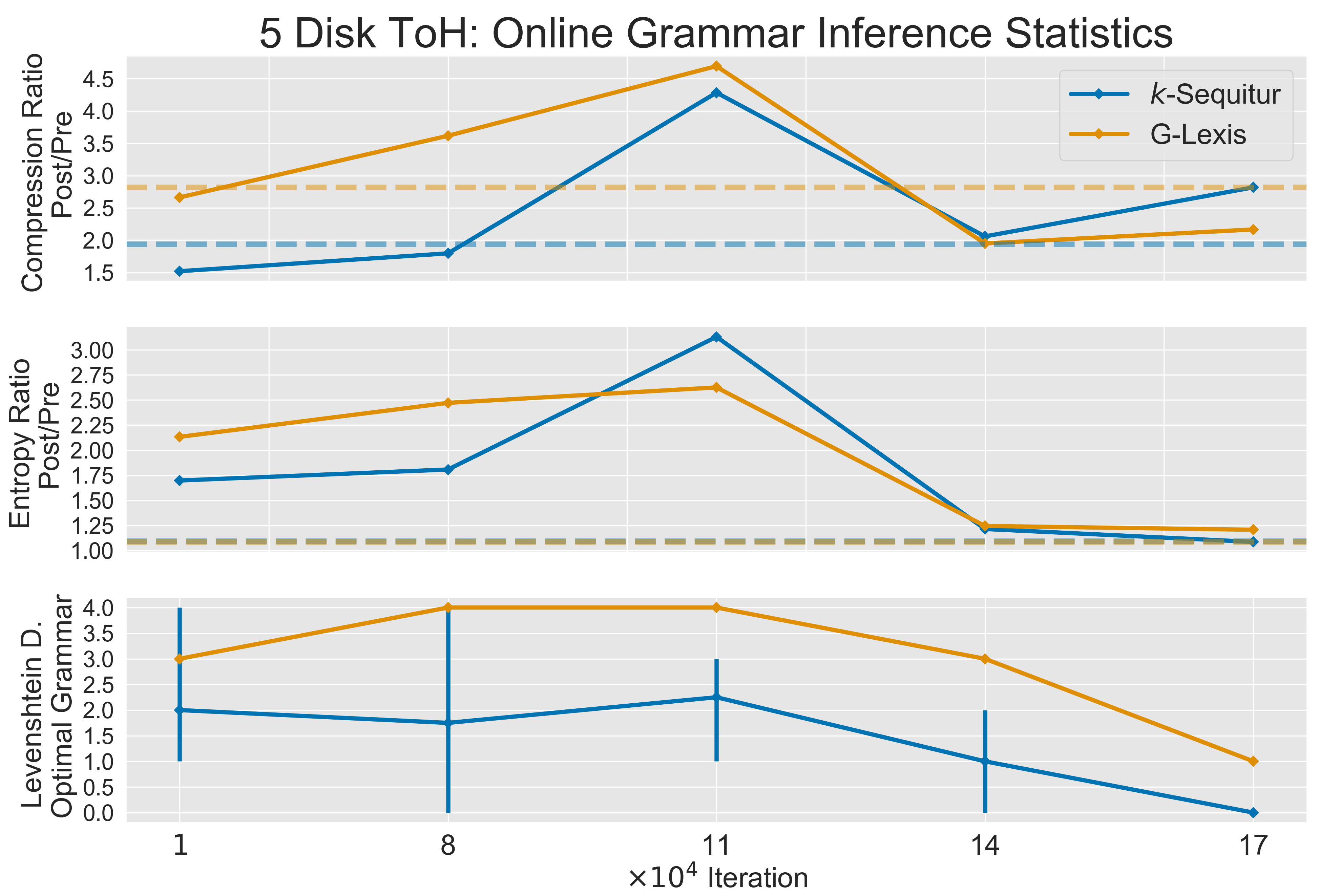}
\endminipage\hfill
\caption{\textbf{Left.} Visualization of the Grammar Macro "B" for the 5 disk ToH problem. \textbf{Right.} Grammar Inference Compression and Convergence Statistics. Horizontal lines correspond to optimal (policy rollout) compression ($|\vartheta|/|\vartheta^{enc}|$) and entropy ratios ($\mathcal{H}(\vartheta^{enc})/\mathcal{H}(\vartheta)$).}
\label{fig:grammar_macro}
\end{figure}

A crucial advantage of the incorporation of a symbolic method such as grammar induction lies in their interpretability. 
The 5-disk Towers of Hanoi flattened production rule $B \to CEd \to bafbcd$ is visualized in figure \ref{fig:grammar_macro} (left-hand side) and captures the recursive nature learned by the grammar. $C \to baf$ moves two disks on the auxiliary pole, while $E \to bc$ moves a third disk from source to target pole and one disk back onto the source pole.
The right-hand side of figure \ref{fig:grammar_macro} shows how different grammar compression statistics evolve over time for the online inferred 5-disk ToH problem. After an initial increase in compression as well as entropy ratios between $\vartheta$ and $\vartheta^{enc}$, both are reduced signalling convergence. The final row displays the Levenshtein string distance distribution between the top 5 inferred productions and the productions corresponding to the grammar inferred from an optimal policy rollout. Again, as the agent learns to solve the task, the inferred grammar and the corresponding set of macro-actions converge to the optimal grammar.

\section{Conclusion}

Inspired by parse trees of sequential behavior and the hierarchical nature of complex skill acquisition, we introduced a novel decision making framework which exploits grammatical inference to identify temporally-extended actions. Our contributions can be summarized as follows: \\
(1) Context-free action grammar agents provide efficient and interpretable solutions to imitation and transfer learning problems. One is able to encode task-specific knowledge and to transfer it to more demanding settings. Formal grammars can distill skills into temporally-extended actions. \\
(2) Alternating between grammar updates and learning action values is an effective method to learn an optimal grammar as well as an optimal policy in an online fashion.\\
(3) Formal grammars inferred on a discrete set of primitive actions yield interpretable temporally-extended actions. The proposed action grammars framework provides a fusion between symbolic and gradient-based learning. \\
Our work expands on two key areas of RL research, Symbolic RL and Hierarchical RL: We extend the ideas of  symbolic manipulation in symbolic RL \citep{Garnelo_2016, Garnelo_2019} to the dynamics of actions as empirically manifested in their sequential execution. Moreover, while Relational RL approaches \citep{Zambaldi_2018} draw on the complex logic-based framework of inductive programming, we merely observe invariants of successful behavioral sequences to induce higher order structure. These hierarchical structures offer not only compact task representations, but also lend themselves to human interpretation, thus tackling RL explainability.
Ultimately, we envision a universal language of action sequences for complex tasks, which provides an expandable library of skills for agents which act in diverse natural environments (e.g. dexterous manipulation, driving, etc).


\newpage

\setlength{\bibsep}{3pt plus 0.3ex}

\bibliographystyle{ecta}
{\bibliography{HRL.bib}}

\newpage
\appendix
\section{Supplementary Material}

In the following section we briefly give an overview of hyperparameters and robustness of results displayed in this report. Code to replicate all the displayed results may be found here: \url{https://github.com/RobertTLange/action-grammars-hrl}.

\subsection{Towers of Hanoi Experiment Details}

The ToH experiments are run with SMDP-Q-Learning and the following hyperparameters

\fcolorbox{black}[HTML]{E9F0E9}{\parbox{\textwidth}{%
\small{
\begin{center}
\begin{tabular}{ |p{3cm}|p{3cm}||p{3cm}|p{3cm}| }
 \hline
 \multicolumn{4}{|c|}{\textbf{Tabular Action Grammar SMDP-Q-Learning Hyperparameters:}} \\
 \hline
Hyperparameter & Value & Hyperparameter & Value\\
 \hline
 Learning rate $\alpha$ & 0.8 & Discount factor $\gamma$ & 0.95\\ 
  Eligibility Trace $\lambda$ & 0 & Exploration $\epsilon$ & 0.1 \\ 
 \hline
\end{tabular}	
\end{center}
}}}

The $TD(\lambda)$ baseline shares all the hyperparameters apart from the eligibility trace $\lambda$ which is set to $0.1$. We train the agents 300000 (5 disks) and 7000000 (6 disks)

The grammar inference schedule as well as the respective $k$ hyperparameters can be found in figure \ref{fig:online_adaptations}.

\subsection{Gridworld Grammar DQN Experiment Details}

\begin{wrapfigure}{r}{0.5\textwidth}
   \centering
    \includegraphics[width=0.75\linewidth]{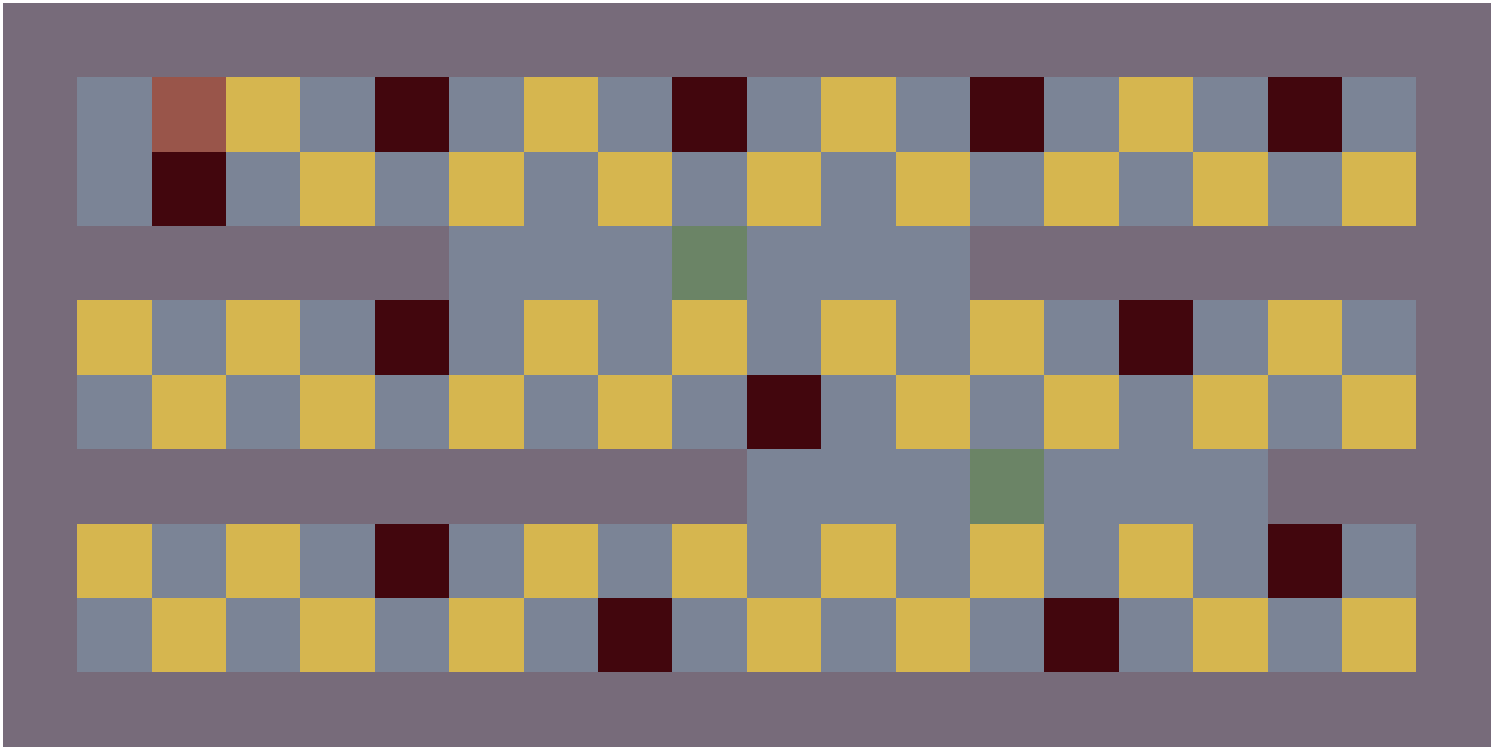}
    \caption{Hierarchically-Structured Grid World Environment.}
    \label{fig:gridworld}
\end{wrapfigure}
 
The gridworld problem provides a non-sparse reward design. The agent (red) has to avoid poisonous items (black) and collect food (yellow). The agent is required to solve a large set of individually smaller sub-tasks. These include 'zick-zack' avoidance of poisonous objects as well as floor traversal. Furthermore, the agent has to avoid terminal collision with the moving blocks (green), whereas the ToH environment rewards the fastest solution. 
The state space is represented by a $(10, 20, 6)$ tuple and the agent has 4 actions available. 
 The fixed architecture of the DQN is a multi-layer perceptron (two-layer 128 hidden units) trained using Adam \cite{Kingma_2014} with a batch-size of 32.

\fcolorbox{black}[HTML]{E9F0E9}{\parbox{\textwidth}{%
\small{
\begin{center}
\begin{tabular}{ |p{3cm}|p{3cm}||p{3cm}|p{3cm}| }
 \hline
 \multicolumn{4}{|c|}{\textbf{Grammar DQN - Multilayer Perceptron Hyperparameters:}} \\
 \hline
Hyperparameter & Value & Hyperparameter & Value\\
 \hline
 Batchsize & 32 & \# Hidden Layer Units & 128\\ 
 Learning Rate & 0.005 & \# Hidden Layers & 2 \\ 
 Momentum & 0.05 & Optimizer & Adam \\
 Discount factor $\gamma$ & 0.99 & & \\
 \hline
\end{tabular}	
\end{center}
}}}

The online Grammar DQN infers a new set of macro-actions every 500 iterations using 2-Sequitur. The action space is augmented with the top-two most used flattened productions in the encoding. 

\end{document}